\documentclass[letterpaper,twocolumn,10pt]{article}
\usepackage{usenix2025}

\usepackage{tikz}
\usepackage{amsmath,amssymb,amsfonts}
\usepackage{algorithm}
\usepackage{algpseudocode}
\usepackage{graphicx}
\usepackage{subcaption}
\usepackage{float}
\usepackage{textcomp}
\usepackage{bmpsize}
\usepackage{xcolor}
\usepackage{rotating}

\begin{document}

\date{}

\title{\Large \bf Privacy-Preserving Model and Preprocessing Verification for Machine Learning}

\author{
{\rm Wenbiao Li}\\
Case Western Reserve University
\and
{\rm Anisa Halimi}\\
IBM Research
\and
{\rm Xiaoqian Jiang}\\
UThealth
\and 
{\rm Jaideep Vaidya}\\
Rutgers University
\and
{\rm Erman Ayday}\\
Case Western Reserve University
} 


\maketitle

\begin{abstract}
This paper presents a framework for privacy-preserving verification of machine learning models, focusing on models trained on sensitive data. Integrating Local Differential Privacy (LDP) with model explanations from LIME and SHAP, our framework enables robust verification without compromising individual privacy. It addresses two key tasks: binary classification, to verify if a target model was trained correctly by applying the appropriate preprocessing steps, and multi-class classification, to identify specific preprocessing errors. Evaluations on three real-world datasets—Diabetes, Adult, and Student Record—demonstrate that while the ML-based approach is particularly effective in binary tasks, the threshold-based method performs comparably in multi-class tasks. Results indicate that although verification accuracy varies across datasets and noise levels, the framework provides effective detection of preprocessing errors, strong privacy guarantees, and practical applicability for safeguarding sensitive data.
\end{abstract}

\section{Introduction}\label{sec:introduction}

Machine learning has revolutionized various industries by enabling systems to learn from data and make intelligent decisions~\cite{Dixon_undated-hy,Jayatilake2021-pz,Wasserbacher2022-wt,Yue2024-zz}. From healthcare diagnostics to financial forecasting, the impact of machine learning is profound and widespread. However, the success of machine learning models hinges not only on the algorithms themselves but also on the quality of the data they are trained on. 

A crucial step in ensuring data quality is preprocessing~\cite{Iliou2015-vw,Alam2019-ag,Garcia2015-zb}. Preprocessing is the stage where raw data is transformed into a clean and usable format. This step includes a variety of tasks such as handling missing values, normalizing data, encoding categorical variables, and removing duplicates. The primary purpose of preprocessing is to enhance the performance of the machine learning model by ensuring that the input data is accurate, consistent, and lacks errors.

The preprocessing stage is susceptible to variations, which can significantly impact the resulting model. For instance, failing to normalize numerical data might cause the model to be dominated by attributes with larger magnitudes, leading to biased predictions. Similarly, overlooking the removal of duplicates can skew results.
These variations can occur due to factors such as different domain expertise, interpretations, or miscommunication among team members handling the data pipeline. 
In this work, we consider 6 frequently used preprocessing steps in total, two of them are required before training a model and the remaining four are additional work and decided by the user whether the following study needs one. More details are in~\ref{sec:preprocessing}
When improper preprocessing choices are made, models may become less reliable, even if the overall performance metrics appear unaffected noticeably. In critical applications, such as healthcare or finance, these inconsistencies can have serious consequences.
Thus, before a model is shared with other users or deployed as a service, it is essential to verify that it has been correctly trained and functions as intended. This verification process helps prevent the propagation of issues that could arise from improper preprocessing approaches, ultimately protecting end-users and the broader systems that rely on these models.

However, verifying a machine learning model is challenging, especially when it comes to confirming the properness of its preprocessing steps, as the personalized and dataset-dependent nature of preprocessing makes improperness difficult to detect. 
Although some of these improperness may not cause a significant change in overall accuracy, they can still introduce biases, reduce the model’s generalizability, or lead to incorrect predictions in specific cases. Even when detailed performance metrics, such as those derived from a confusion matrix, are considered, the verification results can still be misleading. This is because these metrics are highly dependent on the distribution of the testing set, which may not fully represent the diversity or characteristics of real-world data or a specific user's dataset. Consequently, subtle biases introduced during preprocessing might go unnoticed, leading to a model that performs well on paper but fails to deliver reliable or fair outcomes in practice.

To address this limitation, we propose leveraging prediction explanations to evaluate model behavior. Unlike traditional performance metrics, prediction explanations are not affected by the distribution of the testing set. They provide insight into the model's decision-making process, highlighting the features or patterns that influence its predictions. This approach allows us to verify the model’s behavior in a more transparent and unbiased manner (see Table~\ref{tab:motivation_table}).

\begin{table*}[ht]
\centering
\fontsize{10}{12}\selectfont
\begin{tabular}{|l|p{6.5cm}|p{6.5cm}|}
\hline
\textbf{Criteria} & \textbf{Explanations} & \textbf{Traditional Performance Metrics} \\ \hline
\textbf{Interpretability} & Intuitive & Requires understanding multiple metrics \\ \hline
\textbf{Dependence on Testing Set} & Less dependent on test set & Highly dependent on test set \\ \hline
\textbf{Bias Detection} & Identifies subtle biases & May miss biases if test set is unbalanced \\ \hline
\textbf{Privacy} & No need for true labels or balanced test set & Requires true labels and balanced test set \\ \hline
\textbf{Implementation Complexity} & Simple;  & Complex; involves multiple metrics \\ \hline
\textbf{Generalizability} & Reflects diverse model behavior & Limited to test set distribution \\ \hline
\end{tabular}
\caption{Comparison of Explanations vs. Performance Metrics}
\label{tab:motivation_table}
\end{table*}

Even more challenging than verifying preprocessing steps is doing so in a privacy-preserving manner. When dealing with sensitive datasets, such as those containing personal or confidential information, directly sharing data for verification purposes is not an option. This limitation requires the use of specialized techniques that can ensure verification without compromising data privacy.

Privacy-preserving verification techniques are designed to address this challenge by allowing the verification of a model's preprocessing steps and overall integrity without exposing sensitive data. These techniques, such as differential privacy, ensure that the verification process itself does not accidentally leak any private information. For instance, by adding controlled noise to the shared data, differential privacy may allow one to check the properness of preprocessing steps and model training while maintaining a strong privacy guarantee. However, balancing the need for accurate verification with the requirement to protect privacy is a delicate task. Too much privacy protection can obscure critical details necessary for thorough verification, while too little protection can expose sensitive data. 

In this paper, we develop a comprehensive framework to verify the claims made by researchers (or machine learning as a service provider) about models trained on sensitive training datasets. 
In recent work, Halimi et al.~\cite{halimi2022privacy} propose an efficient privacy-preserving scheme to provide verifiability of association studies. Their approach involves the researcher providing a partial noisy dataset (achieving local differential privacy) alongside GWAS results and workflow metadata. The verifier uses a noisy dataset and metadata to replicate the study, and then compares the deviations in results with those obtained from a publicly available dataset. By assessing statistical consistency, the verifier determines the correctness of the published results while ensuring that the additional data provided does not increase privacy risks for participants. The problem we consider in this paper is more challenging, since providing verifiability of an ML model is more complicated than the statistical test in~\cite{halimi2022privacy}. 

Our proposed approach employs model explainers, such as LIME (the local interpretable model-agnostic explanations algorithm)~\cite{Ribeiro2016-na} and SHAP (SHapley Additive exPlanations)~\cite{Lundberg2017-tt}, differentially private data sharing techniques, and ML classifiers. 
LIME is a commonly used tool to provide explanations of model predictions~\cite{Gabbay2021-sa,Holzinger2022-zt,Visani2020-nv,Hase2020-zy}. 
Interpretability is a desirable attribute in ML models, allowing humans to comprehend and trust the model's predictions. The LIME algorithm addresses the interpretability challenge by approximating complex, ``black-box'' models with simpler, localized linear models for individual predictions. These linear models are easier for humans to interpret and provide valuable insights into the decision-making process of the more complex models. LIME has gained popularity in various fields, including finance and healthcare, for risk assessment and diagnostic purposes~\cite{Ribeiro2016-na}.
Similarly, SHAP also addresses interpretability challenges by providing a more nuanced approach to understanding the contributions of individual features in machine learning predictions~\cite{Li2022-kx,Meng2020-yq,Parsa2020-fa}. While LIME focuses on local interpretability, SHAP values offer both local and global insights. By employing the concept of SHAP values from cooperative game theory, SHAP attributes an average contribution to each feature across all possible combinations, ensuring a fair distribution of impact among features~\cite{Lundberg2017-tt}.
 
Although the core strength of a model explainer lies in its capacity to generate interpretable explanations for complex models, this characteristic can be utilized beyond that: specifically, for assessing the similarity between two ML models by measuring the distances between their interpretable explanations on the same testing case and model predictions. 
Recently, the Zest framework~\cite{Jia2021-ic} showed how to employ LIME to compare ML models. The primary idea is to evaluate the similarity of predictions between two ML models using the locality-sensitive nature of LIME. When models produce predictions for a particular input, LIME generates explanations for the prediction. The difference between these explanations from the two models then serves as a metric for model similarity. 

Our main claim is that an ML model that is generated from a differentially private version of the original sensitive training dataset should be "closer" to the original model, compared with another model that is trained with a different training set. 
We utilize explanations that are provided by LIME to quantify this closeness between any two ML models. 
We further claim that an ML model trained on a differentially private version of the properly preprocessed training dataset is closer to the original ML model, whereas models trained on improperly preprocessed training datasets deviate more significantly from the original model. Therefore, we show that such a differentially private training dataset can be used for verification while preserving the privacy of the records in the original training dataset.

Overall, we develop a framework that allows the verification of ML models using differentially private versions of their training datasets and LIME explanations that are obtained from the original model and the model obtained from the differentially private training dataset. Our verification framework notably utilizes an ML classifier which accurately differentiates a properly applied preprocessing pipeline from an improper one. 

Our comprehensive evaluation across three real-world datasets confirms the effectiveness of our proposed privacy-preserving verification framework. The results show that while the ML-based approach consistently outperforms the threshold-based method in binary classification tasks, the two approaches perform similarly in multi-class tasks. Verification accuracy varies across datasets and noise levels, particularly in multi-class tasks, which present additional challenges. Despite these variations, our framework demonstrates strong privacy guarantees and practical applicability, providing a viable solution for protecting sensitive data while ensuring preprocessing properness and model reliability.



\section{Related Work}\label{sec:related_work}

Recent advancements in privacy-preserving verification have been particularly notable in the fields of machine learning and biometric systems. One approach involves using edge computing for privacy-preserving face verification~\cite{Huang2021-ih}, employing secure nearest neighbor algorithms to protect biometric data. However, this method is mainly designed for face data and relies on a specialized computational environment, which limits its use with different data types and machine learning models.

Similarly, privacy-preserving techniques have been applied to speaker verification~\cite{Rahulamathavan2019-nx}, using secure multiparty computation and encryption to protect speaker data. While effective, these methods add significant computational complexity, which can hinder scalability, especially in large-scale applications.

In neural networks, the pvCNN framework~\cite{Weng2023-wt} offers privacy-preserving and verifiable testing specifically for convolutional neural networks (CNNs). Although it ensures privacy and verification, its focus on CNNs limits its applicability to other types of machine learning models, reducing its flexibility.

Additionally, privacy-preserving data integrity verification methods have been developed for mobile edge computing environments~\cite{Tong2019-aa}. These methods, while robust, are specifically designed for edge computing, limiting their usefulness in other contexts. Another approach uses multi-authority attribute-based encryption for verifying smart grid data. Although effective in smart grids, this method’s complexity makes it challenging to apply more broadly.

The ability to compare machine learning models in a privacy-preserving manner is crucial for tasks like model verification and unlearning. A recent method, Zest, introduced by Jia et al.~\cite{Jia2021-ic}, leverages LIME to approximate the global behavior of black-box models through a collection of linear models. Zest calculates model similarity by measuring the cosine distance between the concatenated weights of these linear models, enabling an architecture-independent comparison. This method has proven effective for detecting model reuse and verifying machine unlearning.

Our work leverages the Zest approach of comparing models using LIME explanations, but we focus on a different application: verifying whether a target model has undergone correct preprocessing. While Zest aims to detect model reuse and unlearning across architectures by comparing their functional behaviors, our method applies LIME explanations to ensure that the target model adheres to specific preprocessing steps, without requiring knowledge of the model's true labels. This makes our approach particularly suited for privacy-preserving verification, ensuring that preprocessing integrity is maintained without exposing sensitive data. Unlike Zest, which measures model similarity, we focus on verifying the properness of preprocessing applied to the training data, thereby ensuring the validity of the model under different privacy constraints.

In contrast to these methods, our approach offers a more comprehensive and versatile solution by integrating Local Differential Privacy (LDP) with LIME explanations. By combining LDP with LIME, we provide robust privacy protection while also verifying the model’s behavior in an interpretable and transparent manner. This approach ensures data privacy and extends verification to include preprocessing steps, making it broadly applicable across various machine learning models and data contexts. This versatility and the ability to transparently verify model behavior set our work apart from existing methods and enhance its relevance in real-world scenarios.

\section{Background}\label{sec:background}

In this section, we provide a brief overview of 
differential privacy, Local Interpretable Model-Agnostic Explanations (LIME), SHapley Additive exPlanations (SHAP), and preprocessing in machine learning.

\subsection{Differential Privacy}\label{sec:ldp}
Differential privacy is a mathematical framework designed to protect individual data points when aggregating information across a dataset. Although effective in preserving privacy, it can distort the statistical properties of the data. The privacy level is controlled by the privacy parameter $\epsilon$, which quantifies the trade-off between privacy and data utility~\cite{Dwork2014-fc}.

Local Differential Privacy (LDP) is a variant of this framework that protects individual data by introducing noise at the data point level, ensuring privacy without relying on a trusted data curator~\cite{Cynthia2006-LDP}. Unlike traditional differential privacy, where noise is added to query results, LDP randomizes each data point independently, making it suitable for decentralized data collection environments, such as mobile phones~\cite{Arcolezi2021-jw} and IoT devices~\cite{He2024-dt}.
The effectiveness of LDP hinges on the balance between privacy and utility, which is determined by the amount of noise added. While greater noise enhances privacy by obscuring individual data points, it can also reduce the dataset’s utility. Consistency in noise application is crucial to avoid vulnerabilities, as predictable patterns could compromise the privacy guarantees.

In our work, the researcher (data provider) applies LDP mechanisms to the dataset before sharing it with the verifier (client), ensuring privacy protection without exposing raw data. This one-to-one interaction leverages the core advantage of LDP: protecting data before it leaves the provider's control. By sharing a noisy dataset, we allow the verifier to validate the results of the machine learning model without needing access to the raw dataset, aligning with the decentralized privacy-preserving ethos of LDP.

Achieving LDP commonly involves the addition of noise, with the Laplacian mechanism being a prominent method due to its effectiveness in ensuring privacy while retaining a measure of utility~\cite{Dwork2014-fc}. This mechanism adjusts the amount of noise based on the sensitivity of the function $f$ being applied to the data and the chosen privacy budget $\epsilon$:

\begin{equation}
F(x) = f(x) + Lap\left({\frac {s}{\epsilon}}\right),
\label{eg:laplace}
\end{equation}

where $s$ is the sensitivity of the function $f$. 
In our work, we use the $l_{1}$ sensitivity to add Laplacian noise~\cite{Dwork2014-fc}. 
$Lap(\lambda)$, where $\lambda = \frac{s}{\epsilon}$, denotes sampling from a Laplace distribution with scale $\lambda$ and with a probability density function:

\begin{equation}
\textit{PDF}(x\mid \mu ,\lambda)={\frac  {1}{2\lambda}}\exp \left(-{\frac{|x-\mu |}{\lambda}}\right)\,
\label{eg:density}
\end{equation}
 
where $\mu$ is a location parameter ($\mu=0$ to have a symmetric distribution).

\subsection{Local Interpretable Model-agnostic Explanations (LIME)}\label{sec:lime}
LIME stands out for its capability to offer interpretability for complex machine learning models~\cite{Ribeiro2016-na}. The primary idea of LIME is its unique optimization goal, which seeks to identify an interpretable model \( g(z) \), often a linear model, that best approximates a given complex model \( f(x) \) in the proximity of a specific instance \( x \).

Formally, LIME solves the following optimization problem:

\begin{equation}
\underset{g \in \mathcal{G}}{\text{argmin}} \ \mathcal{L}(f, g, \pi_x) + \Omega(g), 
\end{equation}

where \( \mathcal{G} \) is the space of interpretable models; \( \mathcal{L} \) is a loss function that measures the fidelity of \( g \) in approximating \( f \) locally around \( x \). It is often represented by the mean squared error between the two models over perturbed samples; \( \pi_x \) is a proximity measure indicating the closeness of samples to the instance \( x \); and \( \Omega(g) \) is a complexity penalty on \( g \) to ensure it remains interpretable. This typically limits the number of non-zero coefficients in \( g \).

To provide explanations for a sample $x$, LIME generates a set of perturbed samples around \( x \), uses \( f \) to obtain predictions for these samples, and then trains an interpretable model \( g \) to approximate the behavior of \( f \) in this local region. The explanation \( E \) provided by LIME for the instance \( x \) is derived from the coefficients of the interpretable model \( g \), revealing the contribution of each feature to the prediction made by \( f \) on \( x \). 

\begin{figure}[ht!]
    \centering
    \includegraphics[width=0.9\columnwidth]{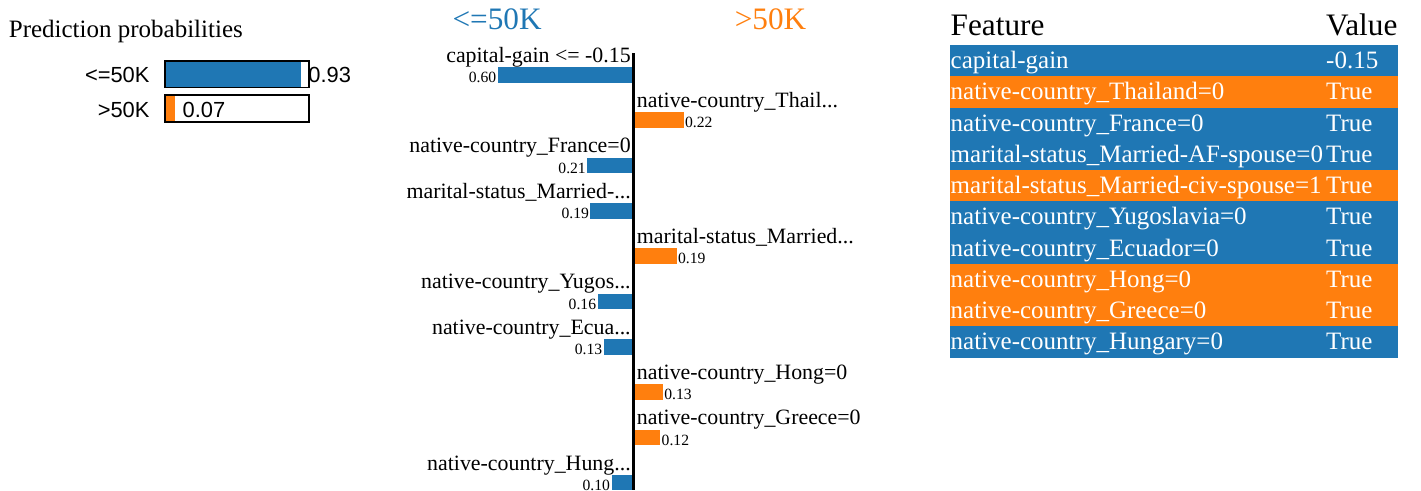}
    \caption{LIME explanations from the Adult dataset~\cite{misc_adult_2}.}
    \label{fig:lime_output}
\end{figure}

We present an illustrative example to demonstrate the type of information that LIME provides for a machine learning model. 
Figure~\ref{fig:lime_output} shows the LIME output of an individual who earns less than or equal to 50K for a logistic regression model trained on the Adult dataset. 
On the left-hand side, the prediction probabilities for each label are presented. In the middle part of the figure, the features contributing to the model's decision are listed. The features are sorted in decreasing order of their weights (explanatory strength). On the right-hand side, the features and their values are listed, color coded to show whether they strengthen or weaken the model's prediction. Further details about the Adult dataset can be found in Section~\ref{sec:evaluation}. 


\subsection{SHapley Additive exPlanations (SHAP)}\label{sec:shap}
SHAP is a method that provides interpretable explanations for the predictions of any machine learning model, emphasizing consistency and accuracy~\cite{Lundberg2017-tt}. It is grounded in the principles of cooperative game theory and utilizes Shapley values to attribute the contribution of each feature to the prediction of a particular instance.

Formally, SHAP can be characterized by the following representation:\

\begin{equation}
\phi(f) = \sum_{i=1}^{N} \phi_i(f, x) = f(x) - f_0,
\end{equation}

where \( \phi(f) \) represents the explanation model for the prediction \( f(x) \); \( \phi_i(f, x) \) denotes the Shapley value for the \( i \)-th feature, quantifying its contribution to moving the prediction from the baseline \( f_0 \) (the average prediction over the dataset) to \( f(x) \), and \( N \) is the number of features.

To provide explanations for an instance \( x \), SHAP values are calculated for each feature, reflecting how much each feature contributes, either positively or negatively, to the prediction \( f(x) \). These values are additive and sum up to the difference between the actual prediction and the baseline prediction, offering a consistent and locally accurate attribution for each feature's effect on the model output.

\begin{figure}[ht!]
    \centering
    \includegraphics[width=0.9\columnwidth]{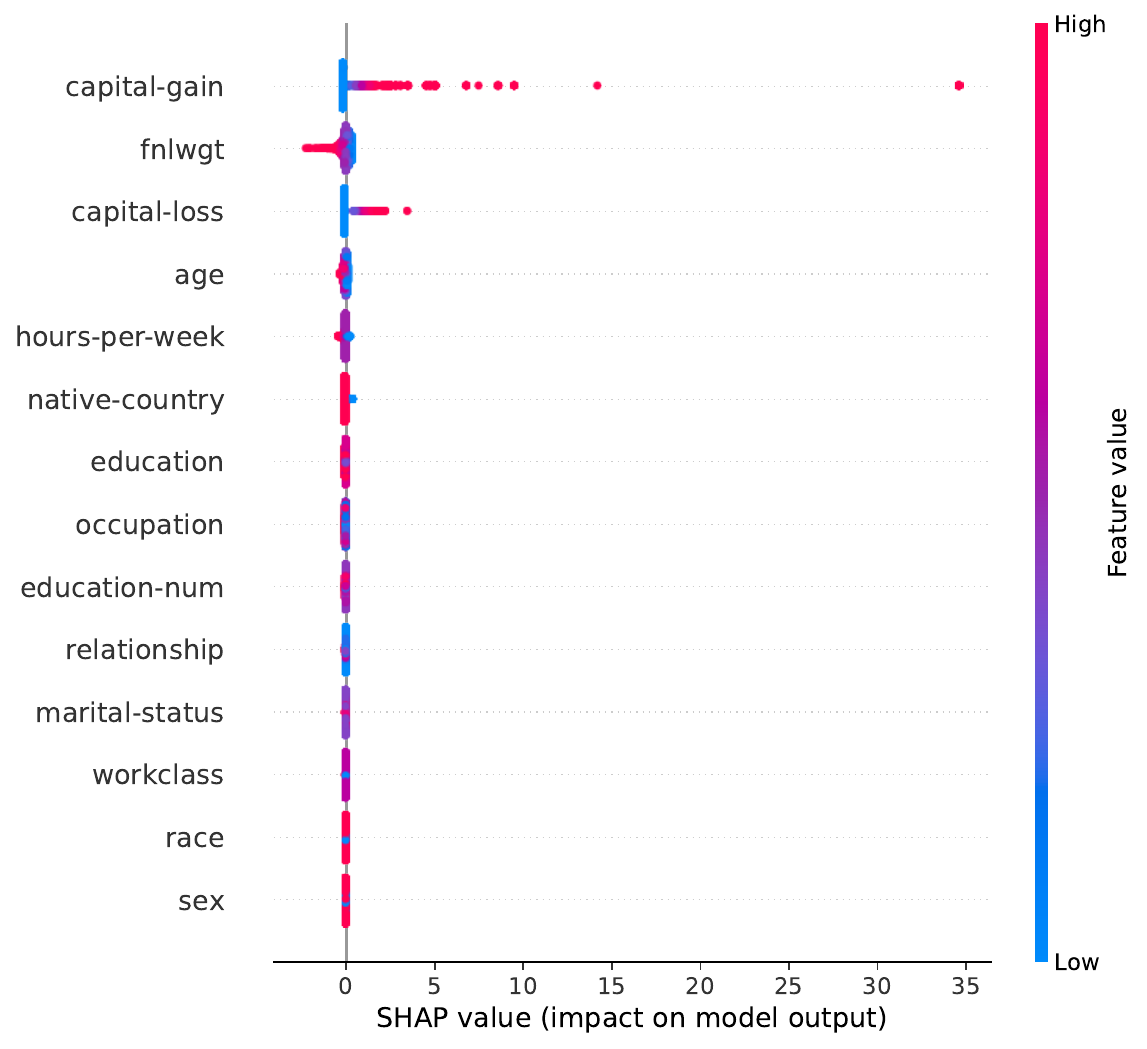}
    \caption{SHAP explanations (Shapley values) from the Adult dataset~\cite{misc_adult_2}.}
    \label{fig:shap_output}
\end{figure}

We also provide an illustrative example shown in Figure~\ref{fig:shap_output} to demonstrate the Shapley values that SHAP provides for the same model and prediction as mentioned above in LIME. The SHAP summary plot organizes features by their importance and impact on the model's output. On the left-hand side, features are listed in descending order of their impact, showing how much each feature drives the model's prediction toward or away from a certain class. The color coding of the values indicates whether a feature's effect increases (red) or decreases (blue) the likelihood of the individual earning more than 50K.

\subsection{Preprocessing in Machine Learning}\label{sec:preprocessing}
Preprocessing~\cite{Garcia2015-zb} is a crucial step in preparing data for machine learning models. Common preprocessing tasks include handling missing values, encoding categorical data, and adjusting for class imbalance, among others. In this work, `dropping missing values' and `encoding non-numeric values' are considered required preprocessing steps. Dropping missing values ensures dataset completeness while encoding non-numeric values into numerical formats, such as using label encoding, allows machine learning algorithms to process the data effectively. These steps are essential for preventing compiling errors during model training.

In addition to the two basic steps, further preprocessing techniques are explored to refine the dataset:

(i) \textbf{Drop Duplicates:} This step involves identifying and removing duplicate rows within the dataset. Duplicate entries can bias the learning process by over-representing certain data points, leading the model to give them undue weight. By eliminating duplicates, each data point contributes equally to the model, thereby enhancing its ability to generalize to unseen data.

(ii) \textbf{Drop Outliers:} Outliers are extreme values that deviate significantly from the rest of the data. This step involves identifying data points that fall beyond three standard deviations from the mean of each feature and removing them to prevent their disproportionate influence on the model. Outliers can skew model results, particularly in algorithms relying on distance measures or means. 

(iii) \textbf{Feature Scaling:} Feature scaling standardizes the range of independent variables to ensure that all features contribute equally to the model. Standardization, which transforms the data so that each feature has a mean of 0 and a standard deviation of 1, is used in this work. This step is particularly important for algorithms sensitive to input data scales. Without feature scaling, features with larger ranges may dominate those with smaller ranges, leading to biased model predictions and suboptimal performance.

(iv) \textbf{Resampling:} Resampling addresses class imbalance by adjusting the dataset to ensure that each class is adequately represented. In cases where one class is significantly underrepresented, oversampling the minority class is applied, creating a more balanced label distribution. This step is crucial in preventing model bias towards the majority class, which could otherwise lead to poor recall or precision for the minority class.

These preprocessing steps serve the purpose of improving data quality and ensuring that the model can learn effectively from the dataset. Proper preprocessing leads to cleaner, more balanced data that allows the model to generalize well and make accurate predictions. On the other hand, improper or incomplete preprocessing can result in biased models, skewed predictions, or overfitting, ultimately diminishing the model’s performance and reliability in real-world applications. Ensuring that preprocessing is done correctly is crucial for developing trustworthy machine learning models that are both accurate and fair.


\section{System and Threat Models}\label{sec:system_threat_model}
In this section, we introduce our system model and highlight potential privacy threats. Our proposed system revolves around two main entities: (i) the researcher, who trains and shares a machine learning model, and (ii) the verifier, who assesses the model's integrity for broader use.

\begin{table}[ht]
\centering
\fontsize{10}{12}\selectfont
\begin{tabular}{|l|p{6.5cm}|}
\hline
\textbf{Symbol} & \textbf{Description} \\ \hline
$D$ & Researcher's original dataset before preprocessing \\ \hline
$D_{\epsilon}$ & Differentially-private version of $D$ \\ \hline
$D_{\epsilon}^{'}$ & Erroneous dataset derived from $D_{\epsilon}$ to represent preprocessing improperness \\ \hline
$D_{\text{test}}$ & Dataset used by the verifier to query the researcher's model \\ \hline
$M_{\text{R}}$ & Researcher's model trained on $D$ \\ \hline
$M_{\epsilon}$ & Verifier's local model trained on $D_{\epsilon}$ \\ \hline
$M_{\epsilon}^{'}$ & Verifier's improper model trained on $D_{\epsilon}^{'}$ \\ \hline
$E$ & Explanations generated by a model explainer ($E^x$ is the explanation for one sample $x$) \\ \hline
$\hat{y}$ & Predicted label ($\hat{y}^x$ is the prediction for a sample $x$) \\ \hline
$O$ & Response pair from a model that includes $E$ and $\hat{y}$ ($O^x$ is the response for one sample $x$) \\ \hline
$V_{\text{ML}}$ & Machine learning-based classifier \\ \hline
$V_{\text{T}}$ & Threshold-based classifier \\ \hline
$\epsilon$ & Privacy budget \\ \hline
\end{tabular}
\caption{Frequently used symbols and notations}
\label{tab:symboltable}
\end{table}

\subsection{System Model}
As illustrated in Figure~\ref{fig:framework}, our framework has two major parties: the researcher and the verifier. The researcher's objective is to construct a machine learning model with high performance (i.e., accuracy) and to make this model accessible to the public. 
To facilitate the verifiability of their ML model, the researcher shares the model architecture (algorithm of the used machine learning model, e.g., logistic regression) without providing its exact trained parameters. The researcher also provides metadata about their training dataset (i.e., a differentially-private version of the training dataset). 
On the other side, the verifier’s goal is to thoroughly assess the integrity of the researcher’s model before it is incorporated into other ongoing studies. In this scenario, the verifier is granted access to critical information, such as the model architecture and the differentially-private original dataset before preprocessing ($D_{\epsilon}$), provided by the researcher. However, the verifier has only black-box access to the model itself. This means they can query the model with data from a testing dataset ($D_{\text{test}}$) and, in return, receive both the model’s predictions ($\hat{y}_\text{R}$) and the corresponding explanations ($E_R$) for each queried data record. These predictions and explanations are used by the verifier to evaluate the model’s behavior, ensuring that it adheres to required standards without revealing sensitive details from the dataset

\subsection{Threat Model}
Here, we outline the potential risks in our framework due to the researcher and the verifier. 

\subsubsection{Researcher}
We consider an honest researcher, that may apply improper preprocessing. We make such an assumption as our framework is not designed to tackle scenarios involving malicious intent. If a researcher intentionally fabricates a training dataset to yield high performance, there is no way to verify this without having access to the training dataset. 
Therefore, our focus is solely on honest researchers who might make improper preprocessing.

Such improper moves may result from improper preprocessing steps. Importantly, they could paradoxically lead to similar or even better performance metrics, such as accuracy, making the improperness hard to detect during development~\cite{Chang2017-uv}. 
This presents a significant challenge, as the researcher might share a model without being aware of the underlying improperness. 
It becomes particularly critical when other researchers (or clients) depend on this model for vital research tasks, potentially unaware of its underlying inaccuracies. 

\subsubsection{Verifier}
The primary goal of our work is to allow the verifier to verify the researcher's model integrity, with high confidence, and in a privacy-preserving way (considering the privacy of the researcher's training dataset). 
However, potential misconduct on the part of the verifier cannot be ignored. Specifically, a verifier might attempt to compromise the privacy of the records in the training dataset by exploiting different attacks such as membership inference, attribute inference, and deanonymization. In the context of membership inference, the verifier aims to discern if a specific record was part of the original training dataset $D$~\cite{Hu2022-fh}. Attribute inference attacks aim to determine undisclosed attributes of entities in the dataset~\cite{Gong2018-dl}, but we consider attributes are unveiled in the differentially-private dataset $D_{\epsilon}$. In deanonymization attacks, the aim is to match anonymized data to some identified data using auxiliary information~\cite{Ding2010-lf}. For instance, this could involve linking anonymized user behavior to individual identities using additional data that hints at who they might be. In our scheme, the researcher shares the training dataset, but it is altered with noise to achieve differential privacy. This setup makes deanonymization attacks quite difficult. An attacker would need much detailed auxiliary information that matches the noisy data. This makes deanonymization more complex than membership inference attacks, where the goal is simply to find out if a particular data record was part of the training set. As outlined by Halimi et al.~\cite{halimi2022privacy}, for the 
considered 
scenario, membership inference becomes the most relevant threat from a dishonest verifier. 
Consequently, in light of the design of our framework and the inherent challenges of successful deanonymization, our primary focus is on addressing membership inference attacks.

Previous studies have shown that membership inference attacks~\cite{Shokri2017-ps, Truex2021-yl, Hu2022-fh} can be effectively mitigated by applying differential privacy (DP) techniques~\cite{Jia2019-xb, Truex2019-ws, Chen2020-ma}. In our approach, noise is added directly to the data to achieve local differential privacy (LDP), ensuring that the sensitive information of any individual is protected while maintaining the overall utility of the shared data. This method provides robust protection against membership inference attacks without the need to alter the dataset by adding or removing records if applying DP. As a result, the power of the membership inference attack due to the shared dataset is significantly reduced, as further discussed in Section\ref{sec:evaluation} of our evaluation."

\section{Proposed Framework}\label{sec:proposed_framework}
In this section, we present our proposed framework for verifying machine learning models, particularly focusing on models claimed to be trained on private datasets.

\begin{figure}[ht]
    \centering
    \includegraphics[width=\columnwidth]{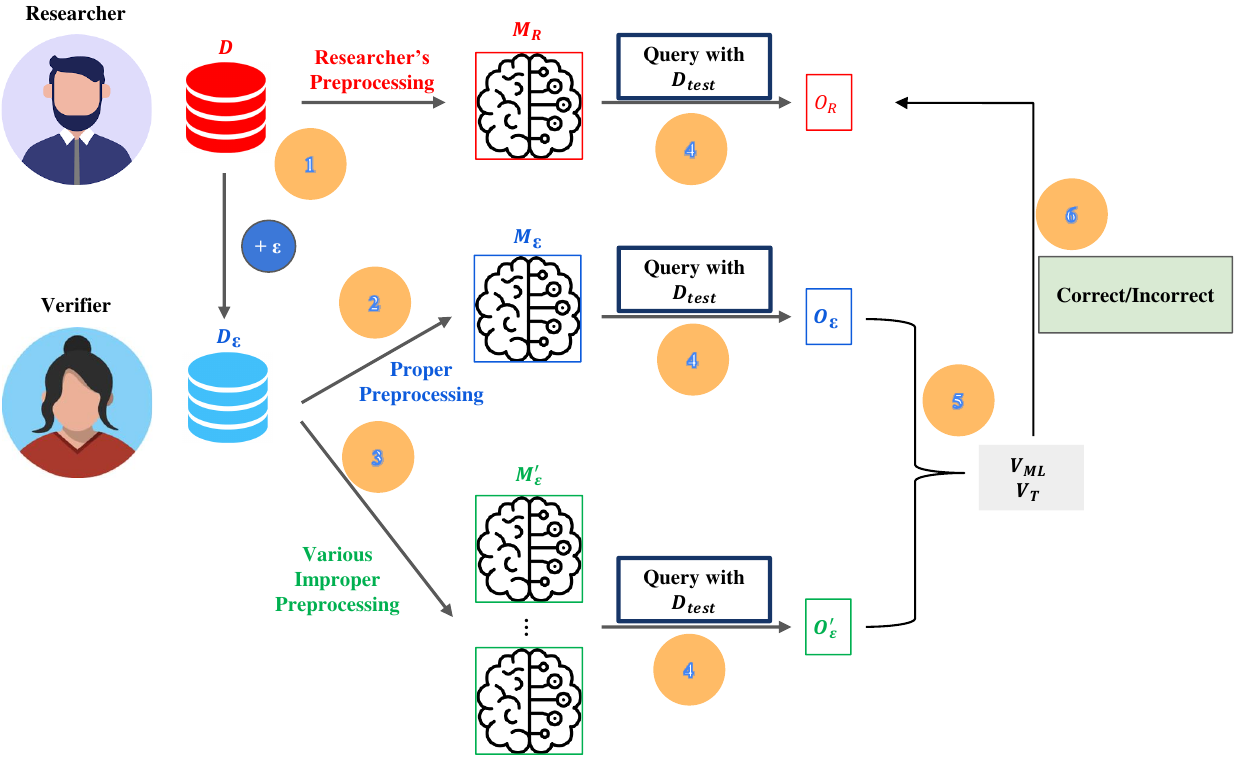} 
    \caption{
    \label{fig:framework}
    \textbf{Overview of the proposed verification framework}. 
    Step 1: The researcher applies improper preprocessing on the original dataset $D$ and trains a machine learning model $M_\text{R}$. 
    Step 2: The verifier accesses the differentially-private original dataset $D_\epsilon$ that is provided by the researcher, applies proper preprocessing, and trains a local model $M_\epsilon$. 
    Meanwhile, in Step 3, the verifier simulates various improper preprocessing to $D_\epsilon$ and trains multiple erroneous models $M_{\epsilon}^{'}$.
    Step 4: In parallel, the verifier sends queries to all trained models from a testing dataset ($D_\text{test}$) and gets their interpretable responses $O$ which contain predictions $\hat{y}$ and explanations ($E$) provided by model explainers (LIME). 
    Step 5: After acquiring response $O_\epsilon$ and $O_{\epsilon}^{'}$ from $M_{\epsilon}$ and $M_{\epsilon}^{'}$, respectively, the verifier builds a machine learning-based classifier ($V_\text{ML}$) and a threshold-based classifier ($V_\text{Th}$). 
    Step 6: The verifier assesses the responses $O_\text{R}$ by checking with the two verifying methods trained in step 5.
    Ultimately, the verifier determines $M_\text{R}$'s correctness.
    }
\end{figure}

\subsection{Steps at the Researcher (Step 1 in Fig.~\ref{fig:framework})}\label{sec:stepsAtResearcher}
The proposed framework identifies distinct roles for the researcher and the verifier. The researcher initiates the process by performing preprocessing to $D$ and trains $M_R$. To safely share the private dataset, the researcher adds Laplacian noise to $D$ to achieve LDP (with parameter $\epsilon$) as discussed in Section~\ref{sec:ldp}. We denote the obtained differentially-private original dataset as $D_{\epsilon}$. 

To create $D_{\epsilon}$, in Alg.~\ref{alg:laplacian_noise}, the researcher processes each feature in $D$ as follows.
It first calculates the sensitivity as the difference between the maximum and minimum values. Given the privacy parameter $\epsilon$, the scale (b) of the Laplacian distribution is determined by the ratio $\frac{sensitivity}{\epsilon}$. Then, it adds noise to each data value, where the noise is randomly drawn from a Laplacian distribution with mean = 0 and the previously calculated $scale = b$. After adding the noise, it rounds (or rescales) the new values within each feature to match the existing possible values and original range of the data.
This method achieves a balance between individual data privacy and the overall dataset utility~\cite{Kairouz2014-jj}. 

\begin{algorithm}
\caption{Generation of Differentially Private Dataset $D_{\epsilon}$ Using Laplacian Noise}
\label{alg:laplacian_noise}
\textbf{Input:} Original dataset $D$, privacy parameter $\epsilon$ \\
\textbf{Output:} Differentially private dataset $D_{\epsilon}$
\begin{enumerate}
    \item \textbf{Initialize:} Set $D_{\epsilon}$ as an empty dataset.
    \item \textbf{For each feature} $F_i$ in $D$:
    \begin{enumerate}
        \item Compute sensitivity: $s = \max(F_i) - \min(F_i)$.
        \item Determine scale of Laplacian noise: $b = \frac{s}{\epsilon}$.
        \item \textbf{For each data point} $x \in F_i$:
        \begin{enumerate}
            \item Sample noise $n \sim \text{Laplace}(0, b)$.
            \item Compute perturbed value: $x' = x + n$.
            \item Post-process $x'$ (e.g., clipping or rescaling) to ensure it remains within the valid range of $F_i$.
        \end{enumerate}
        \item Append the perturbed feature values $F_i'$ to $D_{\epsilon}$.
    \end{enumerate}
    \item \textbf{Return:} Differentially private dataset $D_{\epsilon}$.
\end{enumerate}
\end{algorithm}

When queried by the verifier with $D_{\text{test}}$, in addition to providing the predicted labels $\hat{y}_\text{R}$, $M_\text{R}$ also generates interpretable explanations $E_{\text{R}}$ that include coefficients and intercept of the interpreting linear model. These explanations, $E_{\text{R}}$, are combined with the predicted labels $\hat{y}_\text{R}$ by appending the explanations to the end of the labels to create the final response $O_
\text{R}$.

\subsection{Steps at the Verifier (Step 2 - 6 in Fig.~\ref{fig:framework})}
Recall that the verifier has access to the differentially-private dataset $D_{\epsilon}$ and knows the architecture of the reference model $M_\text{R}$ (e.g., logistic regression).  
Then, the verifier applies both proper and improper preprocessing techniques to $D_{\epsilon}$ and uses the same model architecture to train two types of models: (i) a locally trained model $M_{\epsilon}$ with proper preprocessing; and (ii) one or more erroneous models $M_{\epsilon}^{'}$ with improper preprocessing. Improper preprocessing is detailed further in Section~\ref{sec:evaluation}.

Next, the verifier queries all trained models ($M_\text{R}$, $M_\epsilon$, and $M_{\epsilon}^{'}$) using a testing dataset ($D_\text{test}$) and collects their interpretable responses $O$ ($O_\text{R}$, $O_\epsilon$, and $O_{\epsilon}^{'}$, respectively), and $O = \begin{bmatrix} E & \hat{y} \end{bmatrix} $. $D_\text{test}$ should have the same feature set as $D$. These responses consist of the predictions $\hat{y}$ and the explanations ($E$) generated by a model explainer, such as LIME.

To determine whether $M_\text{R}$ has been correctly trained, which all the needed preprocessing steps have been applied, the verifier employs two approaches: a machine learning-based classifier and a threshold-based classifier.

\textbf{Machine Learning-Based Classifier:} The verifier first creates a training dataset by using the responses $O_\epsilon$ and $O_{\epsilon}^{'}$ as feature values. Labels are then assigned to these responses, resulting in a labeled dataset for training. 
If the verifier seeks to detect specific improper preprocessing steps, this becomes a multi-class classification task, with labels corresponding to each improper preprocessing scenario. Alternatively, if the focus is on simply distinguishing between proper and improper preprocessing, a binary classification task is used, where responses are labeled based on whether the preprocessing was proper.

With the labeled data, the verifier trains a machine learning classifier. After training, this classifier can be used to evaluate $O_\text{R}$, determining whether it indicates any potential improperness in $M_\text{R}$.

\textbf{Threshold-Based Classifier:} The threshold-based approach relies on distance metrics, specifically the cosine distance, to assess response similarity. Similar to~\cite{Jia2021-ic}, the verifier calculates the cosine distance between $O_\epsilon$ and $O_{\epsilon}^{'}$, where $\text{Cosine Distance} = 1 - \frac{O_\epsilon \cdot O_{\epsilon}^{'}}{\|O_\epsilon\| \|O_{\epsilon}^{'}\|}$, and labels these distances in the same manner as the responses in the machine learning-based approach.

Once the threshold-based classifier is trained, the verifier uses it to evaluate the cosine distance between $O_\epsilon$ and $O_\text{R}$. If the distance exceeds a predefined threshold, this signals a potential improperness in $M_\text{R}$, and the verifier reports it as such. The threshold is determined by analyzing the frequencies of the cosine distances obtained from $O_\epsilon$ and $O_{\epsilon}^{'}$, and the mean cosine distance is chosen as the threshold for detection.

\section{Evaluation}\label{sec:evaluation}
In this section, we evaluate our proposed framework using three public datasets. We simulate improper preprocessing, assess verification performance and privacy risk of the data samples in the researcher's dataset.

\subsection{Datasets}
We first use the CDC Diabetes Health Indicators dataset~\cite{kaggle_diabetes_2024}. It is used to predict the onset of diabetes based on diagnostic measurements. The dataset consists of 253,680 instances with 21 features, including attributes such as diet, blood pressure, BMI, and smoking. Each instance is classified as either having diabetes or not.

Then, we select another popular dataset, the Adult dataset~\cite{misc_adult_2}. It is used to predict whether an individual's income exceeds $\$$50K/year based on census data. The dataset contains 48,842 instances and 14 features, including attributes such as age, education, occupation, and hours per week. 

Lastly, we use the Student Record dataset (Predict Students' Dropout and Academic Success)~\cite{predict_students'_dropout_and_academic_success_697}. It contains 4424 instances related to students from various undergraduate programs. The dataset includes 36 features, covering academic paths, demographics, and socioeconomic factors, along with students' academic performance at the end of the first and second semesters. The primary task is to predict whether a student will drop out, remain enrolled, or graduate.

All three datasets involve personal information or health data, which are categories of data that are particularly vulnerable to misuse and require rigorous privacy protections.

\subsection{Improper Preprocessing}
We define proper preprocessing as the application of the required steps: `drop missing values' and `encoding non-numeric values', and all four additional preprocessing steps, applied sequentially in the order of (i) drop duplicates, (ii) drop outliers, (iii) feature scaling, and (iv) resampling. Improper preprocessing, on the other hand, is defined as the omission of one or more of these additional steps, starting from (iv) and moving backward, while always including the required steps. For instance, improper steps when missing one: i, ii iii; improper steps when missing two: i, ii. This approach yields a total of 14 unique combinations of improper preprocessing steps, each representing a different scenario of missing steps. Importantly, the order of the steps is preserved by us, meaning that a step with a lower order is always applied before a step with a higher number.

\subsection{Experiment Setup}
In our experiments, we designed the verification process to evaluate the performance of our approach in detecting if the proper preprocessing steps was applied or any of the preprocessing steps was not applied.
The verification task is set up as either a binary-class classification or a multi-class classification problem. In the binary-class scenario, the verifier's objective is to determine whether any preprocessing steps were not applied to the researcher's data. For the multi-class classification, the verifier must identify the specific preprocessing steps that were mistakenly not applied. This distinction allows us to assess the verifier’s capability in both general improperness detection and more granular improperness classification. 

For all experiments, we initially use logistic regression as the target model architecture. Logistic regression is chosen due to its simplicity and interpretability. we can clearly assess the impact of different preprocessing scenarios on the verification process. Beyond logistic regression, we also test with Decision Tree and Random Forest models, which are interpretable and widely used for various classification tasks. Given that we utilize a model-agnostic explainer, the verification methods are expected to generalize well to these other interpretable models. Thus, while logistic regression serves as our baseline, similar results should be achievable with Decision Tree, Random Forest, and other models that can be similarly interpreted. This assumption is based on the model-agnostic nature of the explainer, which does not rely on the specific architecture of the model being analyzed.

We split each dataset into training and testing sets using an 80:20 ratio. The training data was subjected to the necessary preprocessing steps, including drop missing values and encoding, and option steps from the above sequence. To ensure consistency, if normalization was applied to the training data, the same preprocessing was also applied to the testing set, but done separately to avoid information leakage. From the testing data, we randomly sampled 500 rows to construct $D_\text{test}$, which was used for querying the model and obtaining predictions and explanations. Where applicable, we fixed the random state to ensure the reproducibility of our experiments, and each experiment was repeated 5 times to account for variability in the results.
To evaluate the effects of differential privacy, we tested our framework across various privacy budgets by creating $\epsilon$ values of [0.1, 1, 10, 1000, and $\infty$], representing different levels of privacy guarantees in our analysis.

\subsection{Metrics}
To evaluate the effectiveness of our verification process, we employ two key metrics. The primary metric is ``verification accuracy'', which measures how accurately the verifier can detect improper preprocessing in the target model. 
This metric is essential for both binary-class and multi-class classification tasks, where it reflects the verifier’s ability to correctly identify the presence of improperness and, in the case of multi-class tasks, the specific type of improperness.

\subsection{Verification Results}
Figures~\ref{fig:binary_ML_lime_logres} -~\ref{fig:multi_Th_lime_logres} show the verification accuracy of different classification tasks and models 
for varying noise levels controlled by the privacy budget, $\epsilon$. These figures highlight the performance of our verification approaches using the LIME explainer and Logistic Regression, providing insights into their robustness and effectiveness under privacy-preserving conditions across binary and multi-class settings.

\subsubsection{ML-based Verification Performance}
Binary Classification (Fig.~\ref{fig:binary_ML_lime_logres}) presents verification accuracy for binary tasks using Logistic Regression and LIME across the Diabetes, Adult, and Student Record datasets. The verifier shows strong performance as the privacy budget increases (i.e., higher $\epsilon$ value), with accuracy exceeding 0.8 in most cases. Even under stricter privacy conditions (lower $\epsilon$ value), the method maintains reasonable accuracy, generally above 0.5, despite one case below 0.5 for the Student Record dataset at $\epsilon=1.0$. The overall trend demonstrates that the verifier can effectively identify improper preprocessing steps while the privacy of the researcher's data is preserved.

\begin{figure}[ht]
    \centering
    \includegraphics[width=\columnwidth]{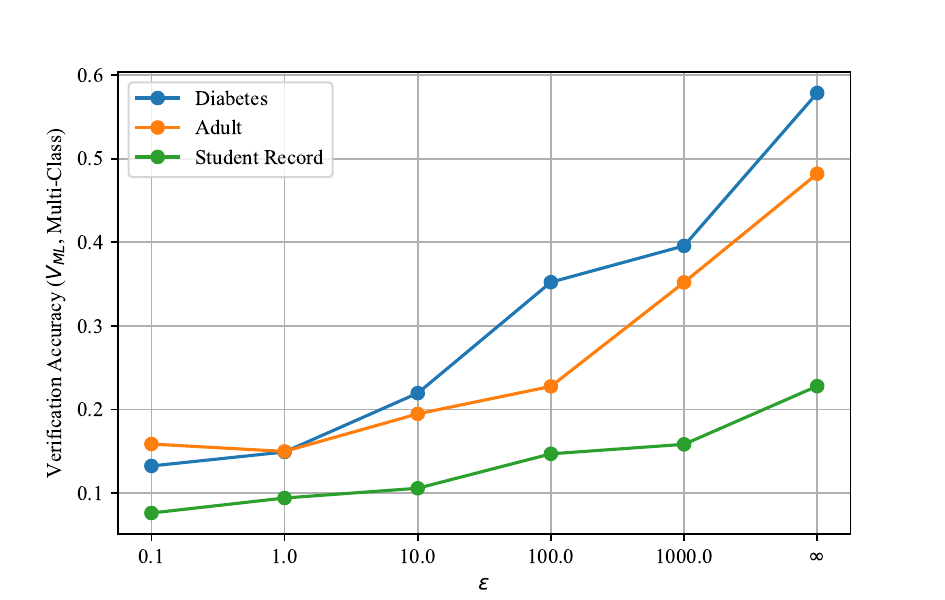}
    \caption{Multi-Class ML-Based Accuracy using LIME explainer on Logistic Regression Model}
    \label{fig:multi_ML_lime_logres}
\end{figure}

Multi-class Classification (Fig.~\ref{fig:multi_ML_lime_logres}) shows the verifier’s performance in handling 15 labels, one for no improperness and 14 for various improper steps. As expected, accuracy is low under a high amount of noise ($\epsilon=0.1$), below 0.2 across all datasets. However, as the amount of noise added decreases (i.e., $\epsilon$ increases), the verifier's performance improves steadily. At higher values of $\epsilon$, accuracy reaches up to 0.6 for the Diabetes dataset and 0.5 for the Adult dataset, though performance is lower for the Student Record dataset, reflecting the challenge of multi-class classification with added noise.

\subsubsection{Threshold-based Verification Performance}
\begin{figure}[ht]
    \centering
    \includegraphics[width=\columnwidth]{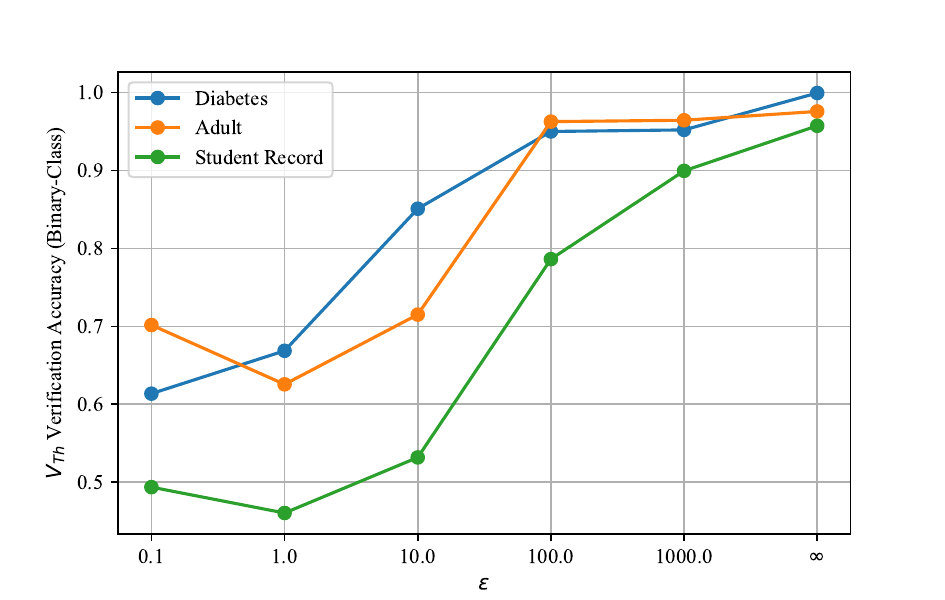}
    \caption{Binary-Class ML-Based Accuracy using LIME explainer on Logistic Regression Model}
    \label{fig:binary_ML_lime_logres}
\end{figure}

\begin{figure}[ht]
    \centering
    \includegraphics[width=\columnwidth]{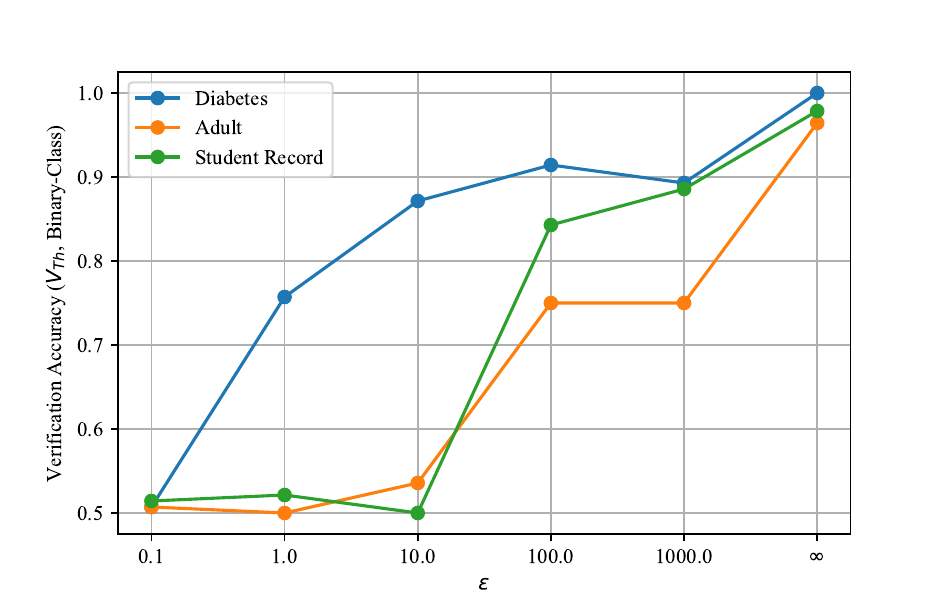}
    \caption{Binary-Class Threshold-Based Accuracy using LIME explainer on Logistic Regression Model}
    \label{fig:binary_Th_lime_logres}
\end{figure}

Binary Classification (Fig.~\ref{fig:binary_Th_lime_logres}) reveals a similar trend to the 
results in Fig.~\ref{fig:binary_ML_lime_logres}, with accuracy increasing as $\epsilon$ increases. The verifier's performance remains above 0.5 for most datasets even at the smallest privacy budgets, and reaches up to 0.9 as the amount of noise added to the researcher's dataset decreases. This demonstrates the threshold-based approach's reliability in managing binary classification under varying noise conditions.

\begin{figure}[ht]
    \centering
    \includegraphics[width=\columnwidth]{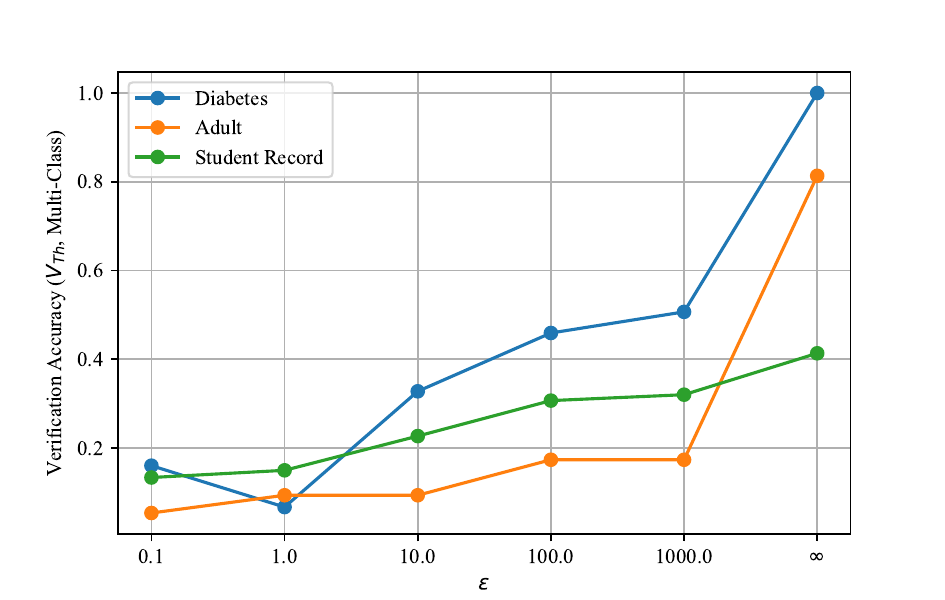}
    \caption{Multi-Class Threshold-Based Accuracy using LIME explainer on Logistic Regression Model}
    \label{fig:multi_Th_lime_logres}
\end{figure}

Multi-class Classification (Fig.~\ref{fig:multi_Th_lime_logres}) shows a more complex scenario, with accuracy starting below 0.2 for all datasets at lower $\epsilon$ values, reflecting the difficulty of distinguishing multiple improper steps under high amount of noise. As the amount of noise decreases, accuracy improves, reaching almost 1.0 for the Diabetes dataset and exceeding 0.8 for the Adult dataset at the highest $\epsilon$ value. While multi-class classification remains more challenging, the threshold-based verifier performs effectively with reduced noise, particularly for these two datasets.

\subsubsection{Additional Model Explainer and Model Architecture}
\begin{figure}[ht]
    \centering
    \includegraphics[width=\columnwidth]{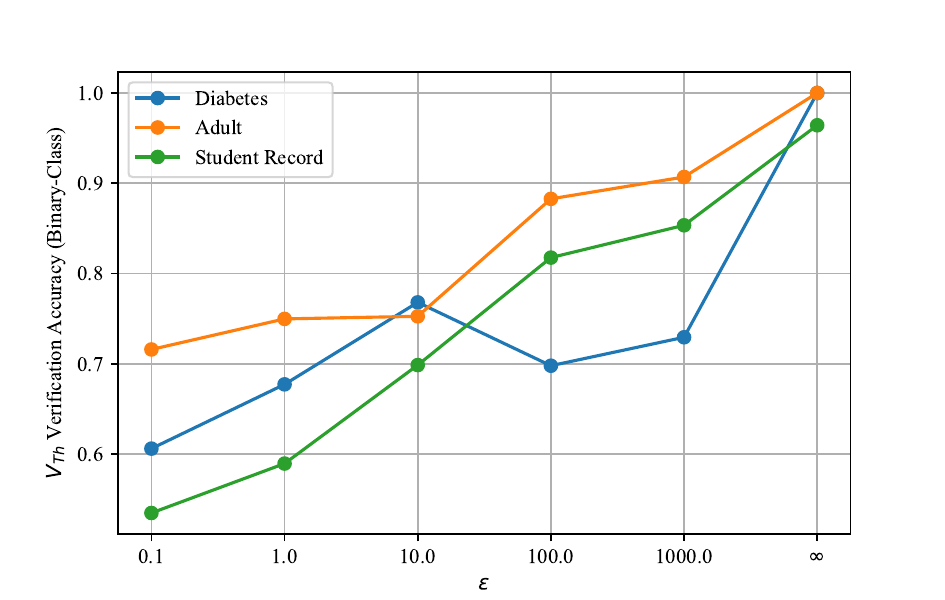}
    \caption{Binary-Class ML-Based Accuracy using SHAP explainer on Logistic Regression Model}
    \label{fig:binary_ML_shap_logres}
\end{figure}

\begin{figure}[ht]
    \centering
    \includegraphics[width=\columnwidth]{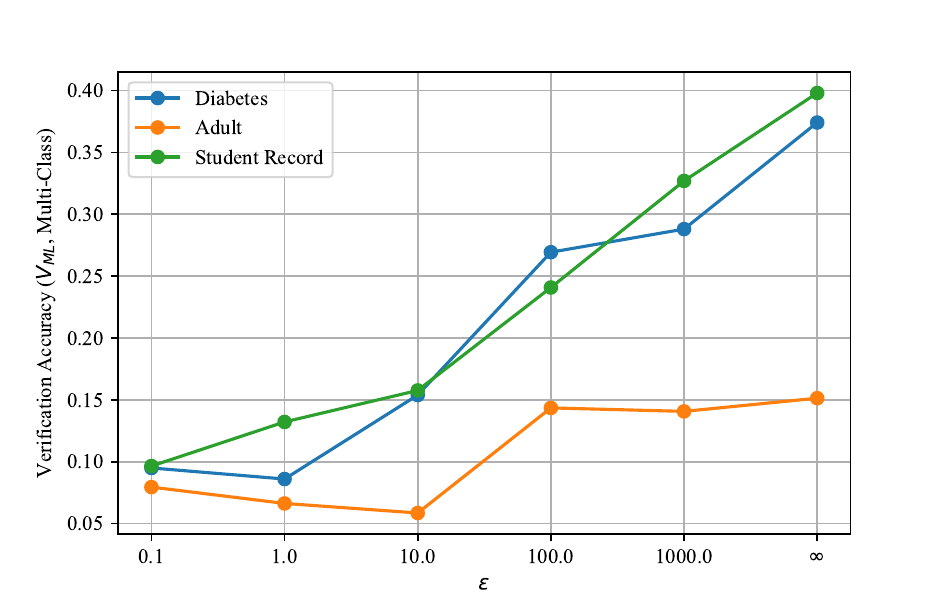}
    \caption{Multi-Class ML-Based Accuracy using SHAP explainer on Logistic Regression Model}
    \label{fig:multi_ML_shap_logres}
\end{figure}

\begin{figure}[ht]
    \centering
    \includegraphics[width=\columnwidth]{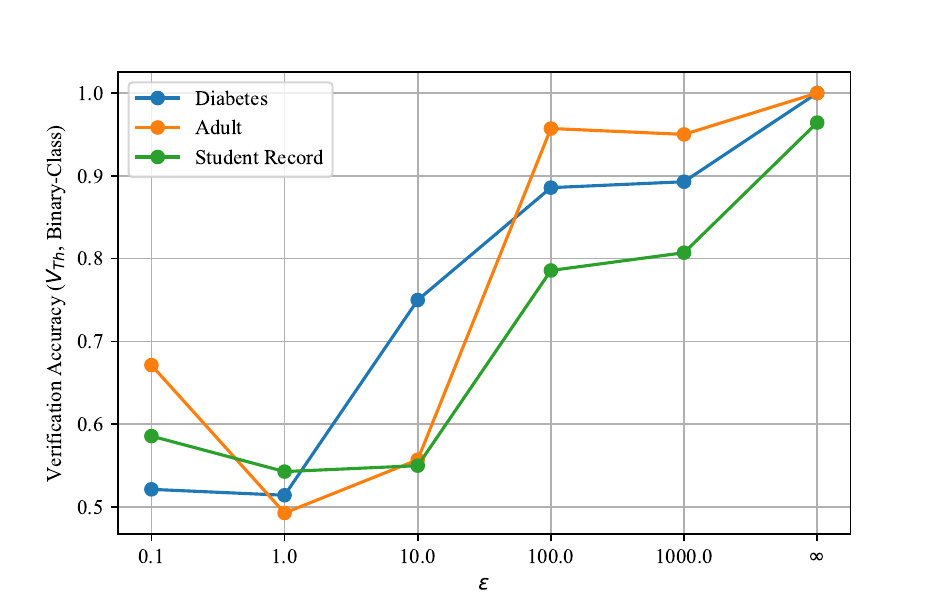}
    \caption{Binary-Class Threshold-Based Accuracy using SHAP explainer on Logistic Regression Model}
    \label{fig:binary_Th_shap_logres}
\end{figure}

\begin{figure}[ht]
    \centering
    \includegraphics[width=\columnwidth]{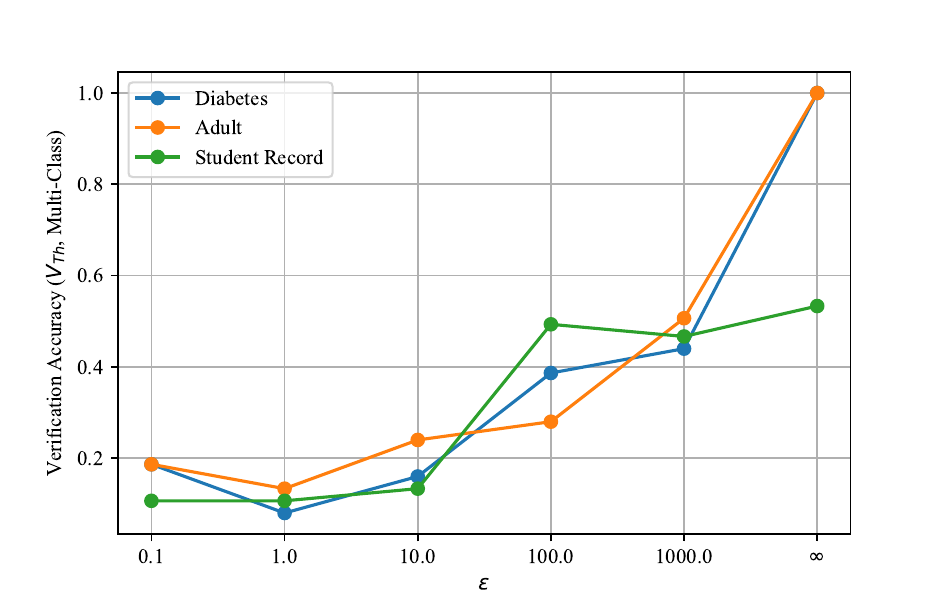}
    \caption{Multi-Class Threshold-Based Accuracy using SHAP explainer on Logistic Regression Model}
    \label{fig:multi_Th_shap_logres}
\end{figure}

We applied the same experimental settings used in previous analyses, with the only variation being the use of the SHAP explainer in place of LIME. By conducting these additional experiments with SHAP (Figs.~\ref{fig:binary_ML_shap_logres} -~\ref{fig:multi_Th_shap_logres}), we aimed to further evaluate the robustness of our framework. The results show that SHAP performs similarly to LIME, maintaining consistent performance in both binary and multi-class classification tasks. For binary classification, the SHAP-based verifier generally achieves high accuracy as the privacy budget increases. In multi-class classification, although the task complexity is higher, verification accuracy still improves as the $\epsilon$ value increases, underscoring the adaptability of the verifier across different explainers.

The inclusion of SHAP strengthens the claim that our framework is versatile and robust across varying privacy settings. The fact that SHAP, alongside LIME, delivers comparable performance further supports the assertion that our approach is not tied to a specific explainer but can work effectively with different explanation techniques. This flexibility is critical in ensuring that the framework remains applicable across diverse use cases while maintaining strong privacy guarantees.

Additionally, we present more results that demonstrate similar findings when using other model architectures, such as Decision Trees and Random Forests (In appendix, Figs.~\ref{fig:binary_ML_lime_dt} through \ref{fig:multi_Th_lime_dt} for Decision Tree, and ~\ref{verifres_rf} for Random Forests). These figures support the general trend observed throughout the evaluation: binary classification consistently achieves higher accuracy than multi-class tasks, especially at higher privacy budgets. By incorporating multiple model architectures and explainers, these results further underline the robustness of our framework, reinforcing its effectiveness as a reliable solution for privacy-preserving verification across different datasets, models, and privacy settings.

\subsection{Privacy Analysis}
In our evaluation, we quantify the power of the membership inference attacks due to the shared differentially private dataset $D_{\epsilon}$.
A misbehaving verifier might attempt to determine whether a target sample is part of the training dataset $D$ used to train the researcher's model $M_\text{R}$ by computing its distance to the differentially-private dataset $D_{\epsilon}$. We refer to this attack as ``Hamming Distance Attack''. 

The verifier might aim to find if any of the samples in the differentially private dataset $D_{\epsilon}$ is a match to a target data sample. To quantify this risk, we use the Hamming distance between the samples. Hamming distance measures the number of positions at which the corresponding elements of two samples differ, making it suitable for categorical or binary data.
First, we create two distinct groups: a case group and a control group. 
The control group contains $|A|$ data samples that are not part of the researcher's dataset $D$ and the case group contains $|B|$ data samples that are part of $D$. 
For each sample $A_x$ in $A$, we compute the Hamming distance between the target sample $A_x$ and all the samples in $D_{\epsilon}$ and identify the minimum Hamming distance. 
Then, we determine the "Hamming distance threshold", $\gamma$, as the $5\%$ false positive rate (i.e., $95\%$ of the samples in $A$ are correctly identified as not being part of $D$). 
Next, for each data sample $B_x$ in $B$, we compute the Hamming distance between the target sample $B_x$ and all the samples in $D_{\epsilon}$ and identify the minimum Hamming distance. 
Finally, we compute the fraction of these $|B|$ data samples that have a minimum Hamming distance that is smaller than the threshold $\gamma$. This fraction gives the membership inference power due to the shared differentially-private dataset $D_{\epsilon}$.

\begin{figure}[ht]
    \centering
    \includegraphics[width=\columnwidth]{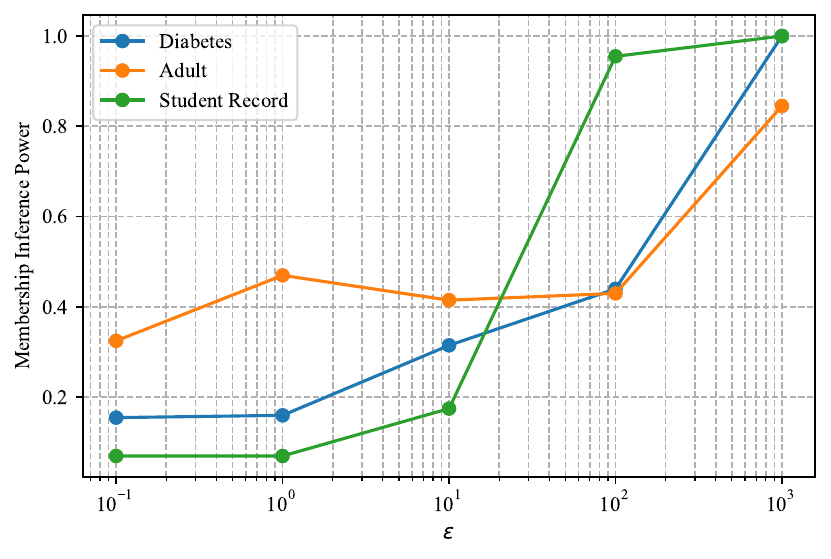}
    \caption{Membership Inference Power via Hamming Distance Attack}
    \label{fig:MIA_power}
\end{figure}

The results are summarized in Fig.~\ref{fig:MIA_power}, where we plot the Membership Inference Power, equivalent to the attacking accuracy, against the privacy budget, $\epsilon$. The privacy budget values tested were $\epsilon = [0.1, 1, 10, 100, 1000]$, and these values are displayed on the x-axis in a log scale to better capture the trend across different scales of privacy protection.

As illustrated in Fig.~\ref{fig:MIA_power}, there is a clear and expected trend: as the privacy budget $\epsilon$ increases, the attacking accuracy or membership inference power also increases. This outcome aligns with our expectations since a higher $\epsilon$ corresponds to less noise added to the shared dataset, making it easier for an attacker to infer membership and, consequently, increasing the success of the membership inference attack. Conversely, when $\epsilon$ is lower, more noise is added, thereby increasing privacy protection and reducing the attack's effectiveness.

An interesting observation from the results is the varying levels of susceptibility to membership inference across different datasets. While the overall trend of increasing attack power with increasing $\epsilon$ holds for all datasets, there are notable differences in the degree of vulnerability. For instance, the Student Record dataset consistently shows a higher membership inference power compared to the Diabetes and Adult datasets at the same $\epsilon$ levels. This suggests that the Student Record dataset is inherently more susceptible to membership inference attacks, possibly due to its underlying data characteristics, which might make it easier for the attacker to distinguish between members and non-members even with added noise.

\section{Discussion}\label{sec:discussion}

In binary classification tasks, the framework consistently maintained high accuracy across all datasets, with minimal fluctuations. Results indicate that verifying whether the model was trained with proper preprocessing is generally easier for the verifier, regardless of dataset complexity. Even at lower privacy budgets (i.e., higher noise), the verifier's performance remained strong, with accuracy frequently surpassing 0.5, demonstrating the robustness of the ML-based approach in simpler tasks.

In contrast, multi-class classification tasks displayed greater variability in performance across datasets. The Diabetes dataset showed relatively stable accuracy, with performance improving as noise decreased (higher $\epsilon$ values). However, the Adult dataset exhibited a broader range of accuracy, and the Student Record dataset experienced the most significant variations. This suggests that dataset complexity, including the nature of the features and the preprocessing steps involved, substantially impacts the verifier's ability to detect improper preprocessing. Simpler datasets, like Diabetes, which contain more homogeneous features, allowed the verifier to perform more consistently. In comparison, more complex datasets like Adult and Student Record presented additional challenges.

Certain types of preprocessing errors were easier to detect in multi-class tasks, particularly those that caused noticeable shifts in data distribution, such as skipping resampling. These significant changes in the dataset led to more pronounced deviations in model behavior, making it easier for the verifier to identify inconsistencies. On the other hand, more subtle preprocessing errors, such as removing duplicates, were harder for the verifier to detect. These changes typically affect only a small subset of features or data points, making them less apparent, especially in datasets like Student Record, which inherently contain more noise and variability.

When comparing the performance of ML-based and threshold-based methods, we found that both approaches performed similarly in multi-class tasks, particularly at higher privacy budgets. Neither method consistently outperformed the other across all datasets, indicating that both can be effective depending on the dataset and the privacy budget. In binary tasks, however, the ML-based approach consistently showed higher accuracy across all datasets. This trend suggests that while both methods are effective for multi-class tasks under the right conditions, the ML-based verifier is more reliable for binary tasks, where the task complexity is lower.

Overall, binary classification consistently achieved higher accuracy than multi-class classification, due to the simpler nature of the binary task. Multi-class tasks, with their increased complexity and number of labels, posed more challenges for the verifier, particularly under higher noise conditions. However, as noise decreased, accuracy improved across both methods and datasets. In multi-class classification tasks, maintaining membership inference power below 0.5 and using the highest $\epsilon$ resulted in comparable performance for both methods.

\subsection{Limitations}
One of the limitations we encountered relates to measuring dataset variation caused by different preprocessing pipelines in order to assess its impact on verification accuracy. While we initially explored using metrics such as Euclidean distance to quantify dataset changes, no clear association between these metrics and the verifier’s performance was found. Although larger dataset variations were expected to correlate with higher verification accuracy, the results across different datasets did not consistently support this hypothesis.
For instance, while the verifier performed well on certain datasets despite moderate dataset variations, other datasets with larger variations did not consistently yield higher accuracy. This suggests that the magnitude of dataset change alone may not fully explain the verifier’s ability to detect improper preprocessing steps. The specific nature of the preprocessing and the complexity of the dataset likely play a more significant role.
Given these findings, we recognize the need for alternative methods to capture and quantify dataset variations more effectively. Future work will focus on identifying metrics that better reflect the structural and semantic changes in the dataset, rather than relying on distance-based measures. This could involve using metrics that consider feature importance, model sensitivity, or other indicators of how preprocessing impacts model behavior.
Moving forward, refining these metrics will be crucial for improving the accuracy and reliability of our verification framework, particularly in complex, multi-class classification tasks. By developing a clearer understanding of how different types of dataset changes influence verification performance, we can make the framework more adaptable and robust across diverse applications.

Another factor contributing to the complexity is the inherent dependency of preprocessing on the dataset. While our experiments used fixed preprocessing steps across all datasets to ensure a controlled environment for assessing verification performance, this approach might not fully reflect real-world situations. In practice, preprocessing pipelines are often tailored to specific datasets after conducting thorough exploratory data analysis and model selection. Our use of fixed preprocessing simplifies these complexities, potentially limiting the applicability of our results. A more adaptive preprocessing strategy, customized for each dataset, could lead to more consistent and meaningful verification outcomes.

\subsection{Future Work}
Several limitations identified in this study provide opportunities for future research. One key area for improvement is the development of more sophisticated metrics to measure the impact of preprocessing. While Euclidean distance offers some insight, it does not fully capture the diverse and complex changes introduced by preprocessing steps. Future work will explore alternative metrics, such as feature importance, model sensitivity analysis, or structural data comparisons, which may better reflect the subtleties of dataset transformations and their impact on verification accuracy.

Another promising direction is the development of adaptive preprocessing strategies. In this study, preprocessing steps were fixed across datasets to maintain consistency, but real-world scenarios often require customized preprocessing based on the dataset's characteristics. Future experiments will implement dynamic preprocessing strategies tailored to each dataset, providing a more realistic and comprehensive evaluation of the verifier’s performance under varying conditions. This would help bridge the gap between controlled experimental settings and practical, real-world applications.

Moreover, future work could focus on extending the scope of verification to include not only model integrity but also model usefulness. For example, verifying whether a pre-trained model underwent appropriate preprocessing tailored to a specific study’s needs could be valuable. If a user is interested in applying a model to a study focusing on a specific demographic, such as middle-aged males with hypertension and diabetes who are smokers, it would be crucial to verify that the model was trained on a dataset filtered for these criteria. Our method could be adapted to check for the presence of such specific preprocessing steps, ensuring that the pre-trained model is relevant and suitable for the study. This approach would enable users to trust that a model meets their specific requirements before deployment.

Finally, future research will explore refining privacy-preserving techniques to address the varying vulnerabilities of different datasets. Our findings suggest that a single privacy budget may not offer uniform protection across datasets. Developing methods to automatically adjust the privacy budget based on dataset characteristics could result in more robust privacy protection while preserving the verifier's effectiveness. This adaptive privacy approach would ensure stronger protection for datasets that are more susceptible to membership inference attacks, like the student record dataset, without compromising performance in less vulnerable datasets.

\section{Conclusion}\label{sec:conclusion}
This paper introduces a privacy-preserving framework for verifying the integrity of machine learning models, focusing on both binary and multi-class verification tasks. In binary tasks, our framework demonstrated that the ML-based approach consistently outperforms the threshold-based method, showing strong accuracy under varied privacy budgets. For multi-class tasks, the ML-based and threshold-based approaches performed similarly, highlighting both methods as viable options across datasets and noise levels.

These findings underscore that our framework is a reliable, adaptable solution for identifying improper preprocessing steps while preserving privacy. Future work will focus on refining verification metrics to better capture dataset variation and exploring adaptive preprocessing strategies. These refinements aim to enhance the framework's applicability in complex real-world scenarios, supporting robust, privacy-preserving verification of machine learning models trained on sensitive data.

\bibliographystyle{plain}
\bibliography{references}

\appendix
\section{Open science}
All datasets used in this study are publicly available. The CDC Diabetes Health Indicators dataset can be accessed via Kaggle~\cite{kaggle_diabetes_2024}, the Adult dataset is hosted on the UCI Machine Learning Repository~\cite{misc_adult_2}, and the Student Record dataset is available through UCI~\cite{predict_students'_dropout_and_academic_success_697}. The code supporting our experiments can be found via the anonymized repository: \url{https://anonymous.4open.science/r/PP-Verification-C1CC}.

\section{Ethical considerations}
In developing our privacy-preserving framework for verifying machine learning models, we have carefully considered the ethical implications of our research, particularly regarding data privacy, fairness, and the responsible use of sensitive information.

Our framework is designed to enhance individuals' privacy and security by employing Local Differential Privacy (LDP) techniques. These techniques ensure that sensitive data used in model verification cannot be traced back to specific individuals. This approach minimizes the risk of exposing personally identifiable information (PII), which is crucial when working with datasets that contain sensitive attributes, such as health data in the Diabetes dataset.

Additionally, we have considered the fairness and equity of our model verification process. The methods we propose are designed to work across various datasets without biasing the results in favor of one specific dataset or feature set. Ensuring that our framework does not disproportionately affect any particular group or type of data is essential to maintaining ethical standards in machine learning research.

While our framework provides strong privacy guarantees, it is important to acknowledge that the effectiveness of privacy-preserving techniques may vary across different datasets, as highlighted in our results. We encourage the responsible use of our framework, ensuring that appropriate privacy budgets are set based on the specific sensitivity of the data and the context in which the verification is being applied.

Lastly, our research is conducted using publicly available datasets (Diabetes, Adult, and Student Record), and no additional collection of private data was necessary. This mitigates the ethical concerns associated with data collection, including the need for informed consent from individuals whose data is used in research.

\section{Verification Results with Decision Tree}\label{verifres_dt}
The following figures showcase additional verification results obtained using the Decision Tree model with the LIME explainer. These results build on the findings presented in the main body, offering deeper insights into the verifier's performance across both binary and multi-class tasks using varied models. While the primary analysis in the main text focused on Logistic Regression, these figures illustrate that the observed trends persist when utilizing a Decision Tree model, further validating the robustness of our framework across different classifiers.

\begin{figure}[ht]
    \centering
    \includegraphics[width=\columnwidth]{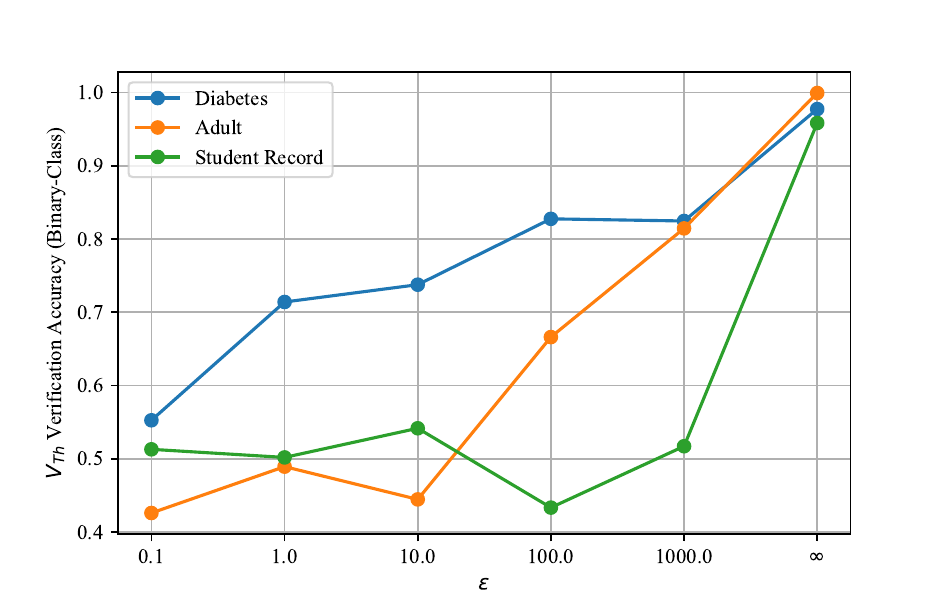}
    \caption{Binary-Class ML-Based Accuracy using LIME explainer on Decision Tree Model}
    \label{fig:binary_ML_lime_dt}
\end{figure}

\begin{figure}[ht]
    \centering
    \includegraphics[width=\columnwidth]{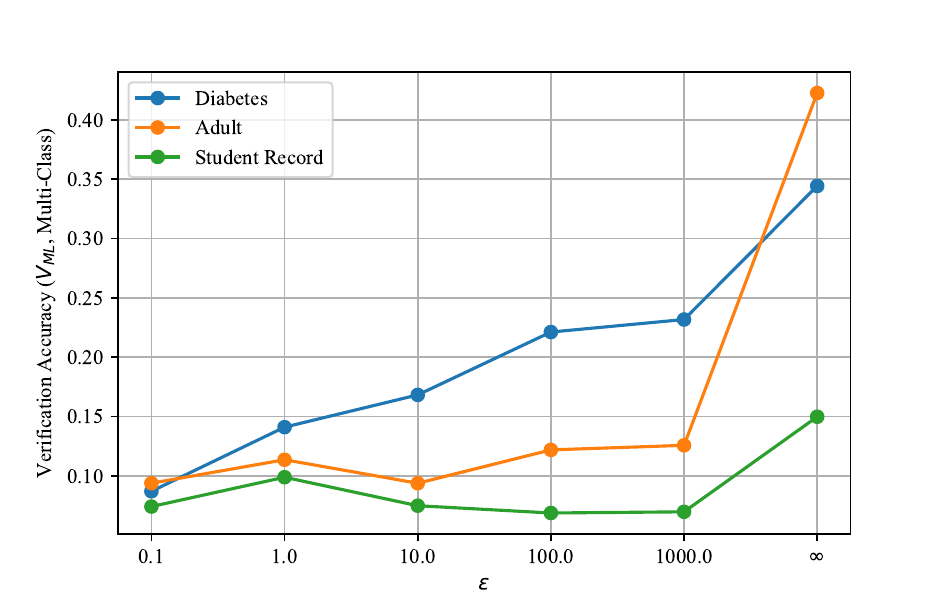}
    \caption{Multi-Class ML-Based Accuracy using LIME explainer on Decision Tree Model}
    \label{fig:multi_ML_lime_dt}
\end{figure}

\begin{figure}[ht]
    \centering
    \includegraphics[width=\columnwidth]{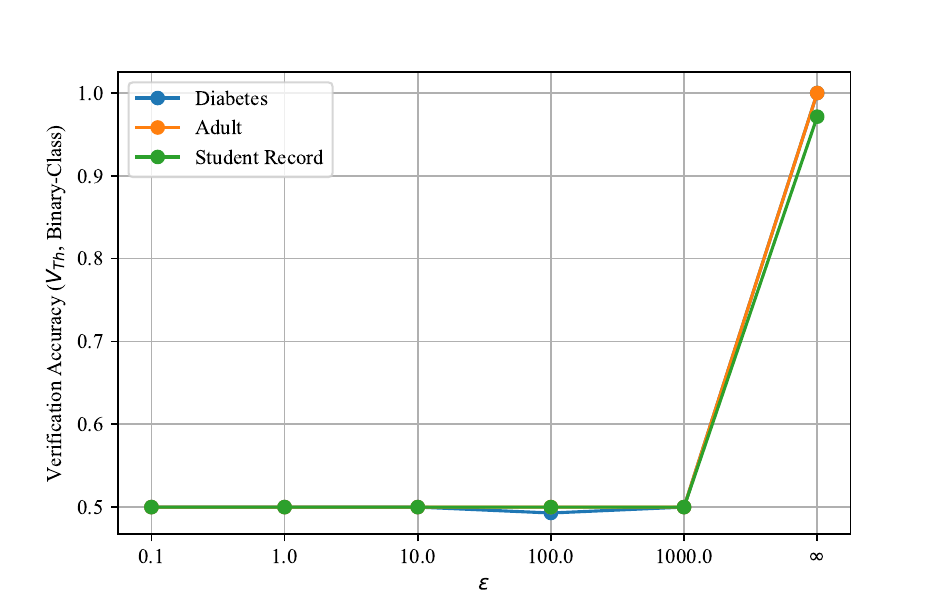}
    \caption{Binary-Class Threshold-Based Accuracy using LIME explainer on Decision Tree Model}
    \label{fig:binary_Th_lime_dt}
\end{figure}

\begin{figure}[ht]
    \centering
    \includegraphics[width=\columnwidth]{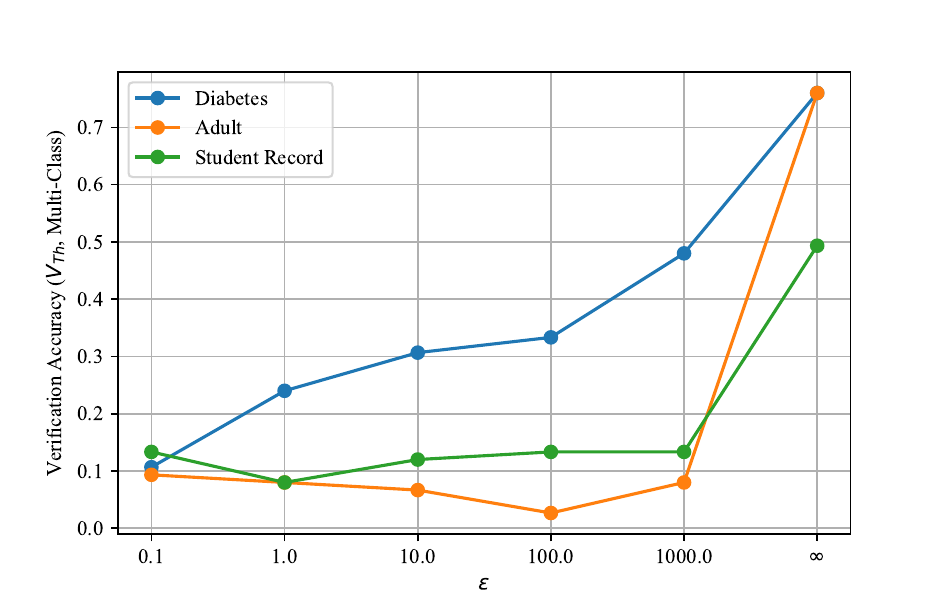}
    \caption{Multi-Class Threshold-Based Accuracy using LIME explainer on Decision Tree Model}
    \label{fig:multi_Th_lime_dt}
\end{figure}

\section{Verification Results with Random Forest}\label{verifres_rf}
The following figures present additional verification results using the Random Forest model with the LIME explainer. These figures complement the results shown in the main body and provide further insights into how the verifier performs across binary and multi-class tasks using different models. While the core analysis in the main text focused on Logistic Regression, these figures demonstrate that the trends observed remain consistent when applying a Random Forest model, reinforcing the robustness of our framework across different classifiers.

\begin{figure}[ht]
    \centering
    \includegraphics[width=\columnwidth]{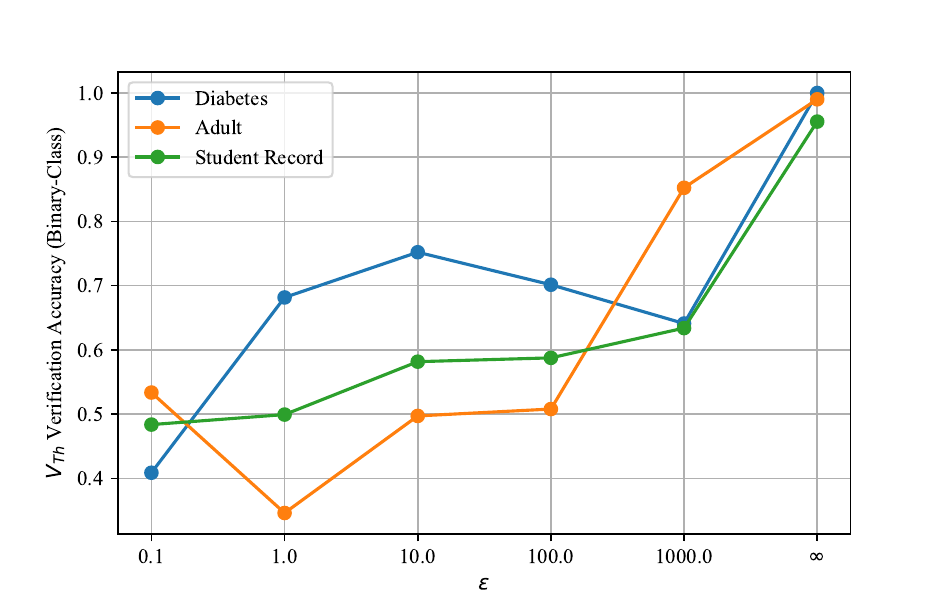}
    \caption{Binary-Class ML-Based Accuracy using LIME explainer on Random Forest Model}
    \label{fig:binary_ML_lime_rf}
\end{figure}

\begin{figure}[ht]
    \centering
    \includegraphics[width=\columnwidth]{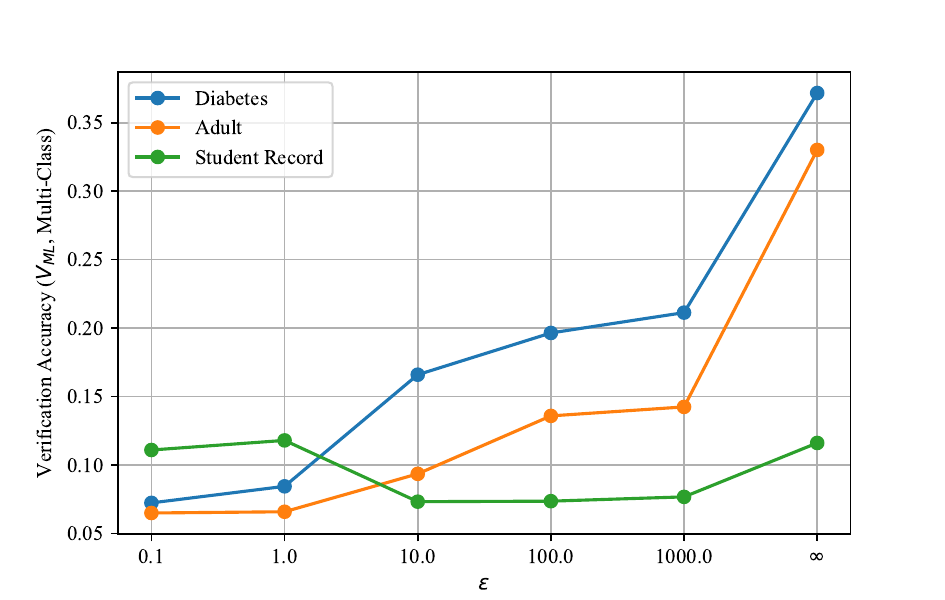}
    \caption{Multi-Class ML-Based Accuracy using LIME explainer on Random Forest Model}
    \label{fig:multi_ML_lime_rf}
\end{figure}

\begin{figure}[ht]
    \centering
    \includegraphics[width=\columnwidth]{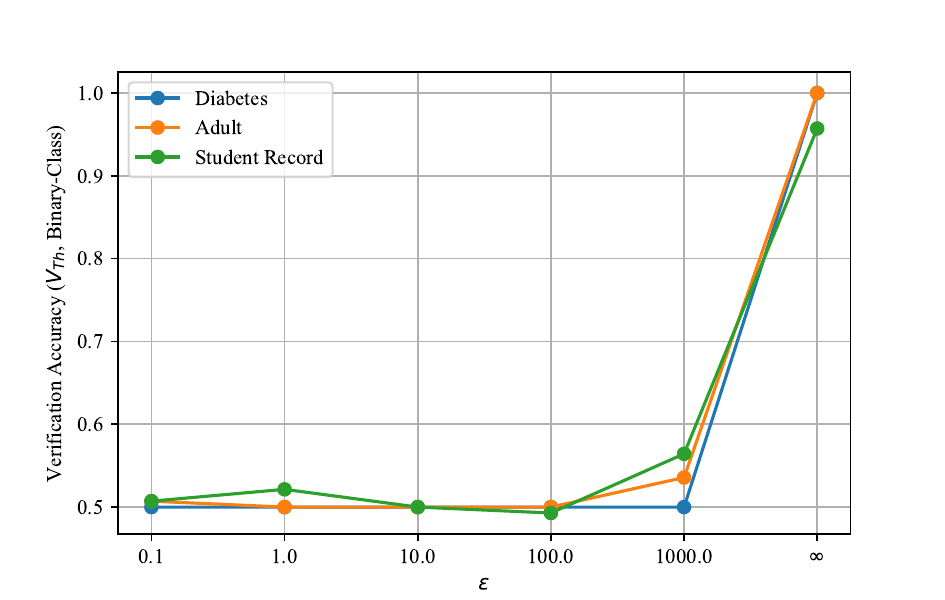}
    \caption{Binary-Class Threshold-Based Accuracy using LIME explainer on Random Forest Model}
    \label{fig:binary_Th_lime_rf}
\end{figure}

\begin{figure}[ht]
    \centering
    \includegraphics[width=\columnwidth]{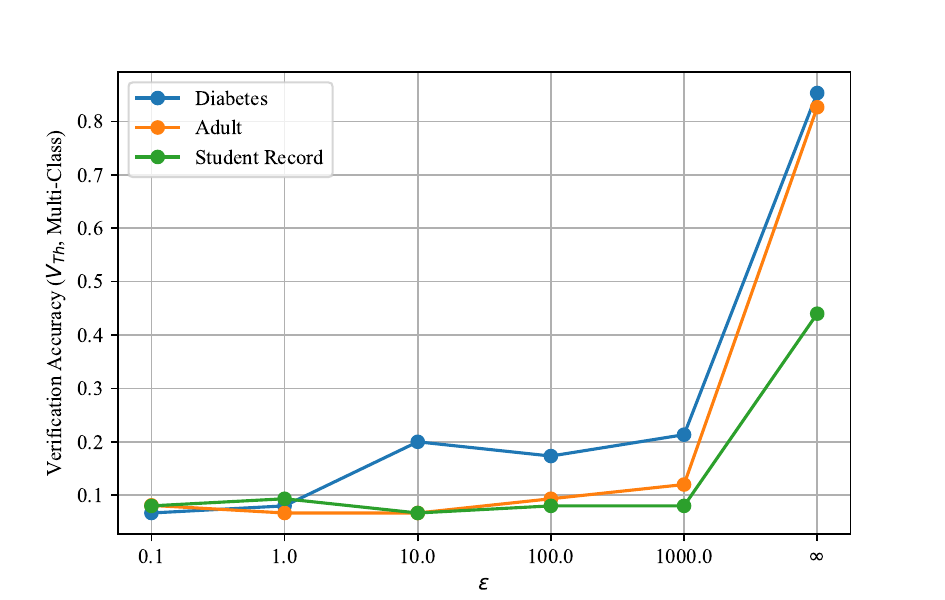}
    \caption{Multi-Class Threshold-Based Accuracy using LIME explainer on Random Forest Model}
    \label{fig:multi_Th_lime_rf}
\end{figure}

\end{document}